\documentclass[letterpaper]{article} 
\usepackage{aaai2026}  
\usepackage{times}  
\usepackage{helvet}  
\usepackage{courier}  
\usepackage[hyphens]{url}  
\usepackage{graphicx} 
\usepackage{natbib}  
\usepackage{caption} 
\usepackage{algorithm}
\usepackage{algorithmic}
\usepackage{amssymb}
\usepackage{amsmath}
\usepackage{multirow}
\usepackage{newfloat}
\usepackage{listings}
\DeclareCaptionStyle{ruled}{labelfont=normalfont,labelsep=colon,strut=off} 
\floatstyle{ruled}
\newfloat{listing}{tb}{lst}{}
\floatname{listing}{Listing}

\title{
   Empowering DINO Representations for Underwater Instance Segmentation \\
   via Aligner and Prompter
}
\author{
    Zhiyang Chen \textsuperscript{\rm 1, \rm 2}\equalcontrib, 
    Chen Zhang \textsuperscript{\rm 1, \rm2} \equalcontrib, 
    Hao Fang\textsuperscript{\rm 1, \rm 2}, 
    Runmin Cong
    \textsuperscript{\rm 1, \rm 2}\thanks{Corresponding author.}
}
\affiliations{
    \textsuperscript{\rm 1}School of Control Science and Engineering, Shandong University, China\\
    \textsuperscript{\rm 2}Key Laboratory of Machine Intelligence and System Control, Ministry of Education, Jinan 250061, China\\

    \{zhiyangchen, chen.zhang, fanghaook\}@mail.sdu.edu.cn, 
    rmcong@sdu.edu.cn
}

\begin{document}
\maketitle

\begin{figure*}[htbp]
    \centering
    \includegraphics[width=500pt]{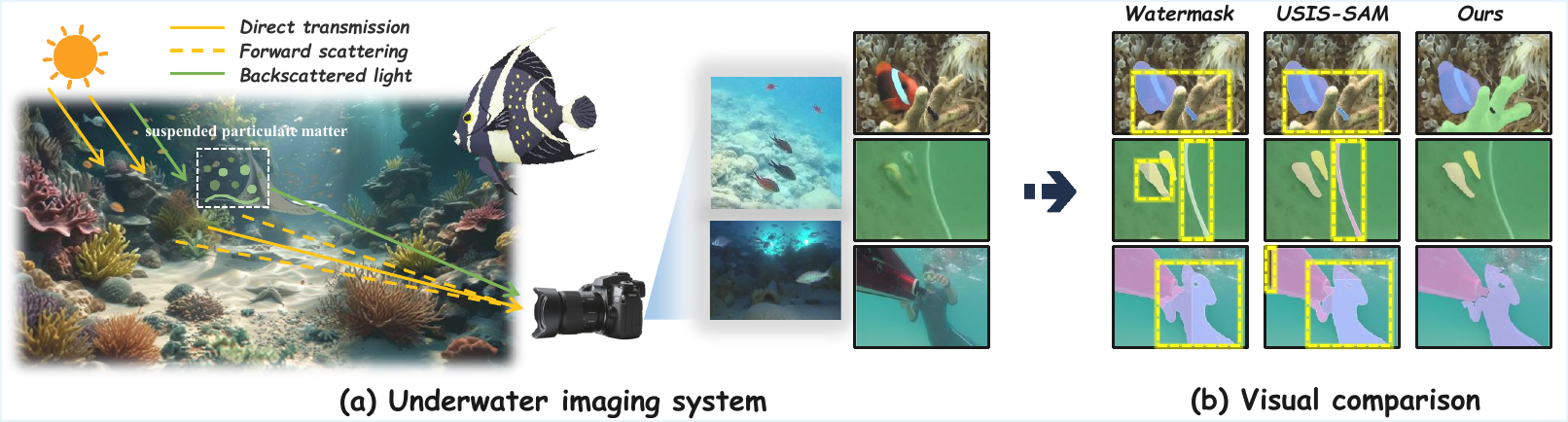} 
    \caption{(a) A typical underwater imaging system: direct transmission carries useful scene information, forward scattering causes blurring, and backscattered light reduces visibility. We also present representative underwater images from UIS datasets. (b) Visual comparisons among \textit{Watermask} (CNN-based method), \textit{USIS-SAM} (SAM-based method), and \textit{Ours} (DINO-based method).}
    \vspace{-3mm}
    \label{fig:fig_1}
\end{figure*}

\begin{abstract}
\textbf{\textit{U}}nderwater \textbf{\textit{I}}nstance \textbf{\textit{S}}egmentation (\textbf{\textit{UIS}}), integrating pixel-level understanding and instance-level discrimination, is a pivotal technology in marine resource exploration and ecological protection. In recent years, large-scale pretrained visual foundation models, exemplified by DINO, have advanced rapidly and demonstrated remarkable performance on complex downstream tasks. In this paper, we demonstrate that DINO can serve as an effective feature learner for \textit{UIS}, and we introduce \textit{\textbf{DiveSeg}}, a novel framework built upon two insightful components: (1) The \textit{AquaStyle Aligner}, designed to embed underwater color style features into the DINO fine-tuning process, facilitating better adaptation to the underwater domain. (2) The \textit{ObjectPrior Prompter}, which incorporates binary segmentation-based prompts to deliver object-level priors, provides essential guidance for instance segmentation task that requires both object- and instance-level reasoning. We conduct thorough experiments on the popular UIIS and USIS10K datasets, and the results show that \textit{DiveSeg} achieves the state-of-the-art performance. Code: https://github.com/ettof/Diveseg.

\end{abstract}

\section{Introduction}
Understanding underwater scenes is critical for advancing marine exploration and sustainable utilization of marine resources, supporting a broad range of applications such as marine research, ecological monitoring, and ocean resource extraction \cite{tang2024neural, abdullah2024caveseg}. 
Among the key technologies in this domain, \textit{\textbf{U}nderwater \textbf{I}nstance \textbf{S}egmentation} (\textit{\textbf{UIS}}) plays a particularly vital role, as it facilitates both pixel-wise classification and instance-level discrimination of underwater objects for accurate recognition and localization. Compared to underwater semantic segmentation \cite{islam2020suim, zheng2024coralscop}, UIS offers significant advantages in distinguishing overlapping objects (\textit{e.g.}, schools of fish and corals) and accurately delineating object boundaries. These capabilities are essential for enhancing the performance of underwater robots or autonomous underwater vehicles in tasks such as obstacle avoidance, object manipulation,  autonomous navigation \cite{cong2021underwater, christensen2022recent} and underwater video segmentation\cite{fang2025decoupled,fang2025learning}.

Unlike natural images, underwater imagery presents unique visual characteristics and significant challenges due to light absorption and scattering, color distortion, low contrast, and limited visibility, all of which substantially degrade image quality  \cite{li2019underwater}. These effects are often non-uniform and depth-dependent, resulting in considerable variation in the appearance of scenes and object instances. A typical underwater imaging system and several degraded images are shown in Figure \ref{fig:fig_1} (a). Early works of UIS \cite{lian2023watermask, jiang2024uw, corrigan2023real} use capacity-limited convolutional neural networks to perform end-to-end learning from collected underwater data. Unfortunately, due to their limited representational capacity, these conventional models still exhibit suboptimal performance in this domain, and some visualization results can be seen in Figure \ref{fig:fig_1} (b). Recent advances in visual foundation models \cite{radford2021learning, kirillov2023segment, oquab2023dinov2,cong2025uis}, trained on large-scale datasets, have inspired growing interest in leveraging pretrained visual embeddings for underwater instance segmentation. Li \textit{et al.} \cite{li2025uwsam} introduced the Segment Anything Model (SAM) \cite{kirillov2023segment} to the UIS task, and proposed using the low-rank fine-tuning technique \cite{hu2022lora} to adapt pretrained representations to underwater scenarios. However, the approach still relies heavily on large-scale underwater datasets to alleviate domain misalignment and yields only marginal performance improvements. 

DINOv2 \cite{oquab2023dinov2} is a powerful visual encoder pretrained in a self-supervised manner on the large-scale LVD-142M dataset, enabling it to learn rich and transferable visual representations. It has shown remarkable generalization ability across diverse downstream tasks, such as object detection \cite{liu2024grounding}, object tracking \cite{tumanyan2024dino}, and image segmentation \cite{li2023mask}. In contrast to SAM, which relies on task-specific supervised training, DINO learns task-agnostic visual features through self-supervised learning. This generalization capability is especially critical in underwater scenarios, where annotated data is scarce and the visual characteristics often deviate significantly from those of natural images. Nonetheless, due to the substantial domain gap, directly transferring DINOv2 to the underwater task is far from straightforward (see Figure \ref{fig:PCA}).

In this paper, we focus on leveraging the DINOv2 foundation model to address the task of underwater instance segmentation, and propose a novel framework, called \textit{\textbf{Diveseg}}. To fully harness the capabilities of DINOv2, it is essential to enhance its adaptation from two perspectives: \textbf{\textit{underwater scene adaptation}} and \textbf{\textit{underwater objects adaptation}}.
(1) Underwater scene adaptation aims to mitigate image degradation caused by color distortion and light scattering effects commonly found in underwater environments. To this end, we design a \textit{\textbf{AquaStyle Aligner}}, which captures the unique stylistic features of underwater images through Fourier decomposition, and then injects them into DINOv2 backbone to eliminate the misalignment with the pretrained model through lightweight parameter learning.
(2) Underwater objects adaptation focuses on enabling the model to generalize effectively to frequently occurring underwater objects—such as corals, jellyfish, and sea turtles—which are underrepresented in the LVD-142M dataset. To tackle this challenge, we introduce the \textit{\textbf{ObjectPrior Prompter}}, which utilizes a binary mask encompassing all foreground objects to facilitate the learning of instance- and class-agnostic features, and subsequently employs them to prompt instance-specific learning. By decoupling instance-agnostic object perception from fine-grained instance discrimination, this mechanism substantially eases the challenge of adapting to diverse marine instances.
We conduct comprehensive experiments on the UIIS and USIS10K datasets, and the results demonstrate that our method achieves superior segmentation performance. The main contributions are summarized as follows:
\begin{itemize}
    \item We are the first to introduce DINO into the UIS task and propose the \textit{DiveSeg}, which effectively eliminates domain inadaptability. Extensive experiments demonstrate that DINO can serve as a powerful learner for UIS and achieves state-of-the-art (SOTA) performance.
    \item We design the \textit{AquaStyle Aligner} to embed the color style features of underwater scenes into the DINO fine-tuning process, enabling better adaptation to underwater environments.
    \item We propose the \textit{ObjectPrior Prompter}, which introduces binary object segmentation cues to provide object-level priors for complex instance segmentation, thereby guiding the model to more effectively localize and distinguish underwater target categories and instances.
\end{itemize}

\section{Related Work}
\subsection{Visual Foundation Models}
In the field of computer vision, visual foundation models have become indispensable for improving downstream task performance. Early approaches such as SimCLR \cite{chen2020simple} and MoCo \cite{he2020momentum} leveraged contrastive learning on tens of millions of images to learn effective feature representations, thereby laying the groundwork for unsupervised visual pretraining. Subsequently, DINO~\cite{caron2021emerging} introduced a self-distillation mechanism to extract robust representations from unlabeled data. Building upon these developments, MAE~\cite{he2022masked} further improved model generalization by performing masked image reconstruction on large-scale datasets. Meanwhile, the SAM \cite{kirillov2023segment} has demonstrated zero‑shot segmentation capabilities for arbitrary visual objects by jointly training on large‑scale manually annotated masks and automatically generated data, thereby introducing a new paradigm for interactive and automated segmentation. 

More recently, DINOv2 \cite{oquab2023dinov2} was pretrained on over one billion web‑sourced images within a multi‑modal processing pipeline, yielding substantial gains in classification, detection, and segmentation tasks.
Unlike SAM, which focuses on precise boundary delineation, DINOv2 captures rich, high-dimensional semantic features that can be seamlessly integrated into downstream segmentation frameworks, thereby enhancing fine-grained semantic understanding in complex scenarios.

\subsection{Underwater Instance Segmentation}
Underwater instance segmentation aims to accurately identify and segment each individual object in underwater scenes, offering both pixel-wise classification and instance-level distinction. Lian \textit{et al.} \cite{lian2023watermask} formally introduced the underwater instance segmentation task by constructing the USIS dataset and proposing WaterMask—an extension of Mask R‑CNN adapted for underwater environments. While WaterMask integrated a dedicated module to mitigate light attenuation and color distortion, its dependence on conventional convolutional backbones limited its representational capacity, leading to suboptimal performance.
To advance the field, Lian \textit{et al.} \cite{lian2024diving} subsequently, the USIS10K benchmark dataset was released-this dataset focuses on salient instance segmentation on the basis of underwater salient object detection~\cite{li2025fscdiff,jin2024underwater}, and improves SAM by introducing USIS-SAM. Through carefully designed adapters and prompt mechanisms, USIS-SAM transfers the pre-trained capabilities of SAM to the underwater domain. However, because the prompt strategy does not fully accommodate the high target density and visual variability that are characteristic of underwater scenes, the segmentation accuracy of USIS-SAM remains limited.

In this work, we propose a novel underwater instance segmentation architecture built upon DINOv2. By harnessing the superior high‑dimensional semantic features of DINOv2 and integrating a lightweight, scenario‑specific aligner, our network substantially outperforms prior methods. Notably, it achieves these gains with only a modest increase in trainable parameters, demonstrating both effectiveness and efficiency in tackling the unique challenges of underwater instance segmentation.

\section{Method}
\subsection{From Motivation to Overall Architecture}
\begin{figure}[tbp]
  \centering
  \includegraphics[width=1\linewidth]{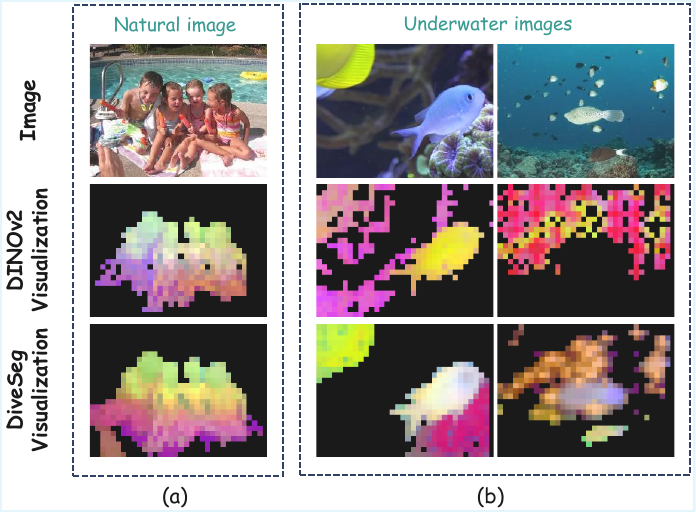}
  \vspace{-4mm}
  \caption{The PCA visualization of DINOv2 and DiveSeg on natural image and underwater images. The background is removed by thresholding the first PCA component.}
  \vspace{-3mm}
  \label{fig:PCA}
\end{figure}

Benefiting from self-supervised pretraining on the large-scale LVD-142M~\cite{oquab2023dinov2} dataset, DINOv2 demonstrates strong general-purpose visual feature extraction capabilities and excels across various downstream tasks. However, in underwater environments, the absorption and scattering of long-wavelength light by water leave primarily short-wavelength components, causing images to appear predominantly blue-green. In Figure \ref{fig:PCA}, we investigate whether the vanilla DINOv2 can extract effective feature representations from underwater images using the principal component analysis (PCA). The results suggest that while DINOv2 is able to capture most of the primary targets in underwater images, its representations are often affected by background noise compared to natural images and may fail to detect some objects.

\begin{figure*}[ht]
    \centering
    \includegraphics[width=\textwidth]{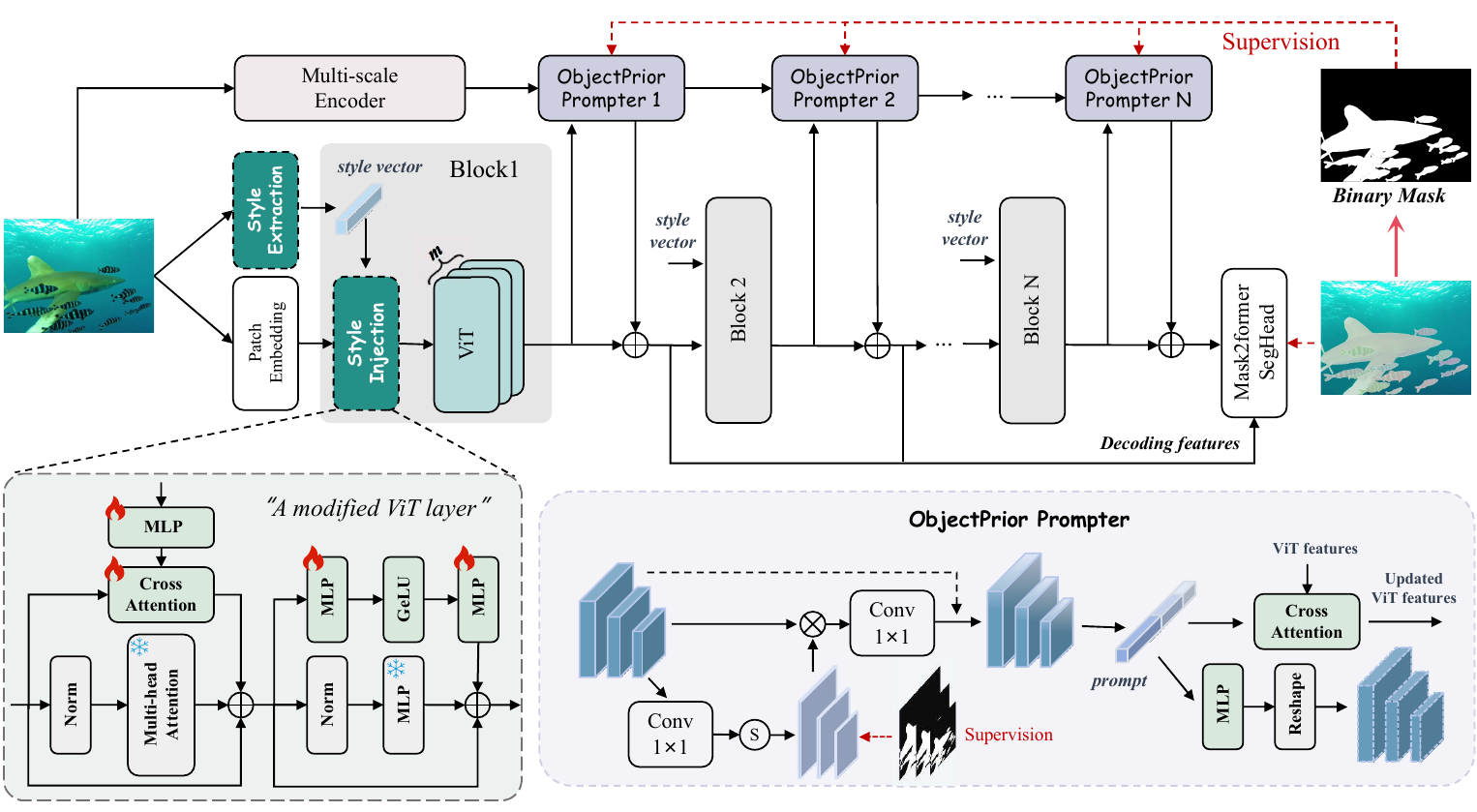} 
    \caption{The overall framework of the proposed \textit{DiveSeg} is illustrated as follows. First, we employ a \textit{Style Extraction} module to obtain an underwater style vector. This vector is subsequently injected into the frozen DINOv2 backbone via the \textit{Style Injection} module, enabling rapid adaptation to the underwater domain. Together, these two modules constitute the AquaStyle Aligner. In addition, the \textit{ObjectPrior Prompter} leverages binary masks to learn object-level priors, which guide the network to focus on underwater objects and ease the challenge of directly segmenting specific instances.}
    \label{fig:example}
    \vspace{-3mm}
\end{figure*}

Hence, to obtain effective representations for UIS, specialized fine-tuning strategies are needed to facilitate the efficient adaptation of DINOv2 to underwater environments. This paper presents \textit{DiveSeg}, a novel framework whose overall architecture is depicted in Figure~\ref{fig:example}, comprising two major components: (1) The \textit{AquaStyle Aligner} addresses the domain adaptation challenge from a scene-level perspective by explicitly extracting background color information to model the unique stylistic features of underwater images, thereby mitigating color domain misalignment; (2) The \textit{ObjectPrior Prompter} tackles the adaptation problem from an object-level perspective by leveraging binary object segmentation masks to guide the learning of target categories and instances in underwater scenes. The two adaptation strategies jointly empower DINOv2 to extract more discriminative and accurate representations tailored to underwater scenes, as shown in Figure \ref{fig:PCA}.


\subsection{AquaStyle Aligner}
\textit{AquaStyle Aligner} contains \textit{Style Extraction} that captures underwater color style information via frequency-domain decomposition, and \textit{Style Injection} that integrates these features into the DINOv2 representation learning process through a cross-attention mechanism.


\begin{figure}[!t]
  \centering
  \includegraphics[width=1\linewidth]{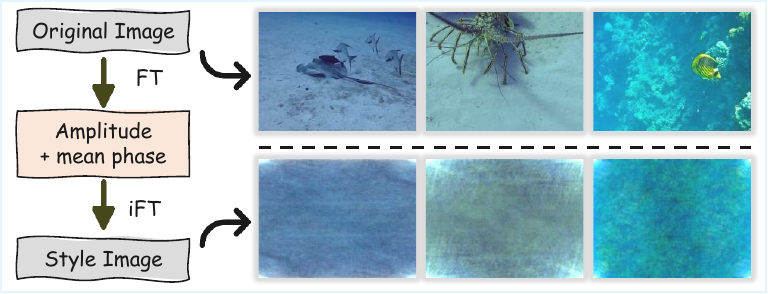}
  \vspace{-4mm}
  \caption{Underwater images and the corresponding style images, FT and iFT represents Fourier transform and inverse Fourier transform.}
  \vspace{-3mm}
  \label{fig:fft}
\end{figure}


\begin{table*}[ht]
    \small 
    \begin{center}
    \renewcommand{\arraystretch}{1.1}
    \setlength{\tabcolsep}{5mm}
    {\begin{tabular}{c|c|c|ccc}
    \hline\hline
    {Method} & {Backbone}   & {Params} & $mAP$ & $AP_{50}$ & $AP_{75}$ \\
    \hline
    \multicolumn{6}{c}{CNN-Based Methods} \\
    \hline
    Mask R-CNN & ResNet-101 & 63M & 23.4 & 40.9 & 25.3 \\
    Mask Scoring R-CNN & ResNet-101 & 79M & 24.6 & 41.9 & 26.5 \\
    Cascade Mask R-CNN & ResNet-101 & 88M & 25.5 & 42.8 & 27.8 \\
    BMask R-CNN & ResNet-101 & 66M & 22.1 & 36.2 & 24.4 \\
    Point Rend & ResNet-101 & 63M & 25.9 & 43.4 & 27.6 \\
    $\mathrm{R}^3$-CNN & ResNet-101 & 77M & 24.9 & 40.5 & 27.8 \\
    Mask Transfiner & ResNet-101 & 63M & 24.6 & 42.1 & 26.0 \\
    Mask2Former & ResNet-101 & 63M & 25.7 & 38.0 & 27.7 \\
    WaterMask & ResNet-101 & 67M & 27.2 & 43.7 & 29.3 \\
    \hline
    \multicolumn{6}{c}{Transformer-Based Methods} \\
    \hline
    USIS-SAM  & ViT-H & 701M & \underline{29.4} & \underline{45.0} & \underline{32.3} \\
    \textit{DiveSeg} (Ours) & ViT-L & 390M & \textbf{35.6} & \textbf{52.0} & \textbf{38.5} \\
    \hline\hline
    \end{tabular}}
    \end{center}
    \vspace{-3mm}
    \caption{Quantitative comparisons with state-of-the-arts on the UIIS dataset. The best results are highlighted in bold, while the second-best results are indicated with underlines.}
    \label{tab:UIIS}
    \vspace{-3mm}
\end{table*}

\textbf{Style Extraction}. A primary distinction between underwater and natural images lies in the color distortion caused by light dispersion in the water medium. This color deviation predominantly manifests in the low-level features of the image. In the frequency domain, the amplitude component of the Fourier spectrum effectively preserves low-level statistical characteristics. Therefore, we adopt the Fourier amplitude as a representation of underwater style features. Specifically, for an image $x\in \mathbb{R}^{H \times W \times 3} $, its Fourier transform $ \mathcal{F}_x $ can be expressed as

\begin{equation}
\mathcal{F}_x(u,v) = \sum_{i=0}^{H-1} \sum_{j=0}^{W-1} x(i,j) \cdot e^{-2\pi {I
} \left( \frac{ui}{H} + \frac{vj}{W} \right)}.
\end{equation}
Here, ${I}$ represents the imaginary unit, and each channel of the image is calculated independently. The amplitude component $A_x$ and the phase component $\phi_x$ are respectively expressed as:
\begin{equation}
	|A_x(u, v)| = \sqrt{\text{Re}\{F_x(u, v)\}^2 + \text{Im}\{F_x(u, v)\}^2},
\end{equation} 

\begin{equation}
	\phi_x(u, v) = \arctan \left(\frac {\text{Im}\{F_x(u, v)\} }{\text{Re}\{F_x(u, v)\}}\right) ,
\end{equation} 
where Re($\cdot$) and Im($\cdot$) are the real part and the imaginary part respectively. To represent the image style information, we fix the phase at the average value $\bar{\phi_x}$ and use the amplitude information to reconstruct the style image $\hat{x}$ via the inverse Fourier transform $\mathcal{F}^{-1}$:

\begin{equation}
    \hat{x}(i,j) = \mathcal{F}^{-1}\!\left\{ |A_x(u, v)| \cdot e^{{I}\bar{\phi_x}} \right\} .
\end{equation} 

As illustrated in Figure~\ref{fig:fft}, averaging the phase while retaining the amplitude in the frequency domain effectively removes object-related content from the image, preserving the distinctive color characteristics of underwater scenes. Based on this, we employ multi-layer convolution and global average pooling to encode the style image into a compact style vector $p_{x}$.


\textbf{Style Injection}. Inspired by the adapter design proposed by Houlsby \textit{et al.} \cite{houlsby2019parameter}, we develop a style injection module that employs cross-attention rather than relying solely on the standard multilayer perceptron (MLP), enabling more effective interaction between style representations and image features. Specifically, implemented as a parallel branch to the Multi-head Attention with normalization layer (MHA) in ViT, our style injection module shares the same input but integrates the style vector $p_x$ through a cross-attention mechanism. In this design, the query is obtained from ViT features, while the key and value are generated from the style vector, enabling effective fusion of visual content and style information. This injection process can be expressed as:
 \begin{equation}
    \omega_1 = MHA({V}_{in}) + {CrossAttn}\left({V}_{in},{MLP}(p_x)\right),
\end{equation} 
where ${V}_{in}$ represents the input features of the ViT block, and all parameters of the \textit{MHA} are frozen. Essentially, the features produced by the cross-attention can be regarded as a complementary representation to those from the original MHA, helping to prevent the degradation of the pretrained model. The output of this step is denoted as $\omega_1$.
 
At the subsequent Feed-Forward (FF) in ViT block, which consists of an MLP followed by a normalization layer, following the standard adapter design paradigm, we employ cascaded multilayer perceptrons to integrate the features from the previous step at a deeper level. This enables DINOv2 to better capture the distinctive color characteristics of underwater images and acquire more discriminative visual representations. This is formulated as:
\begin{equation}
    \omega_2 = FF(\omega_1) + {MLP} \left(GeLU ({MLP}(\omega_1)) \right).
\end{equation} 
Likewise, the parameters of the \textit{FF} remain frozen. Two additional MLP layers are employed to reduce and subsequently restore the feature dimensionality, forming a bottleneck structure. The output of this step, denoted as $\omega_2$, is taken as the output of the ViT block. 

Notably, we evenly divide all ViT layers in the original DINOv2 architecture into four sequential blocks to adapt to the instance segmentation decoder. To minimize the computational overhead and parameter growth, the \textit{AquaStyle Aligner} is only applied to the first ViT layer in each block, resulting in a total of four \textit{AquaStyle Aligners} across the feature extraction process.


\subsection{ObjectPrior Prompter}
Along with the color adaptation for underwater scenes largely resolved, the instance segmentation of underwater-specific objects continues to pose significant challenges. This is primarily due to the rarity of these object categories in the DINOv2 pretraining dataset, rendering direct learning of instance-level segmentation in underwater contexts a nontrivial task. Driven by this insight, we introduce the \textit{ObjectPrior Prompter} that initially applies a binary mask to constrain class-agnostic and instance-agnostic feature learning, subsequently embedding these features as prior prompts within the DINOv2 backbone to ease the learning of instance-level underwater objects. 

As illustrated in the top section of Figure~\ref{fig:example}, we initially extract image features at multiple resolutions through a multi-scale encoder. Subsequently, the \textit{ObjectPrior Prompter} interacts with the backbone to progressively incorporate underwater object priors. Specifically, the multi-scale encoder follows a simple design, consisting of three convolutional layers for feature extraction, stride-2 convolutions for downsampling, and 1×1 convolutions for dimensionality reduction. The encoder outputs a three-scale pyramid of features, denoted as \{$f_{M}^{1},f_{M}^{2},f_{M}^{3}$\}, with corresponding resolutions \{$\frac{1}{8^2}$, $\frac{1}{16^2}$, $\frac{1}{32^2}$\} of the input image. 







In the \textit{ObjectPrior Prompter}, the multi-scale features are first used to generate pseudo masks via 1$\times$1 convolution layers followed by Sigmoid functions, supervised by a binary mask. The binary mask is a single-channel image derived from the segmentation ground truth, covering all target objects. During both training and inference, pseudo masks $P_{mask}$ are dynamically generated for each image to represent all foreground instances. It can be formulated as:
\begin{equation}
    P_{\text{mask}}^{k} = \sigma\left({Conv}_{1 \times 1}^{k}\left( f_{M}^{k} \right) \right),
\end{equation}
where $k \in \{1,2,3\}$. Then, we perform element-wise multiplication between the pseudo masks and the corresponding original features to preserve object-related information. The resulting features are further integrated with the original ones via a convolutional layer and residual connection, enhancing instance-level representations without compromising the original semantics. Formally defined as:
\begin{equation}
   f_{MT}^{k} = Conv_{1 \times1}(P_{mask}^{k} \cdot f_{M}^{k}) + f_{M}^{k},
\end{equation} 




Subsequently, the features $f_{MT}^{k}$ are flattened and concatenated, termed as $O_{prompt}$, and used as an object-level prior prompt to interact with ViT features. In the cross-attention mechanism, the prompt serves as the key and value, while the ViT features act as the query:
\begin{equation}
   f_{ViT}^{opp} = {CrossAttn}\left(f_{ViT}, O_{prompt}\right).
\end{equation} 
Here, $f_{ViT}$ and $f_{ViT}^{opp}$ denote the ViT features before and after prompt-based interaction, respectively.
Finally, we update the features $O_{prompt}$ through multiple linear layers, and then reshape them back to the original pyramid feature dimensions, allowing supervision from the binary mask. We embed the \textit{ObjectPrior Prompter} module after each block and use the sum of its output $f_{ViT}^{opp}$ and the original ViT feature $f_{ViT}$ as the decoder input.

\begin{table*}[ht]
    \small 
    \begin{center}
    \renewcommand{\arraystretch}{1.1}
    \setlength{\tabcolsep}{3mm}
    {\begin{tabular}{c|c|c|ccc|ccc}
    \hline\hline
    \multirow{2}{*}{Method} & \multirow{2}{*}{Backbone}   & \multirow{2}{*}{Params} &\multicolumn{3}{c|}{Class-Agnostic} & \multicolumn{3}{c}{Multi-Class}\\ \cline{4-9}
     &  &   & $mAP$ & $AP_{50}$ & $AP_{75}$ & $mAP$ & $AP_{50}$ & $AP_{75}$\\
    \hline
    \multicolumn{9}{c}{CNN-Based Methods} \\
    \hline
    S4Net  & ResNet-50 & 47M&  32.8 & 64.1 & 27.3 & 23.9 & 43.5 & 24.4 \\
    RDPNet  & ResNet-101 & 66M & 54.7 & 78.3 & 63.0 & 39.3 & 55.9 & 45.4\\
    OQTR & ResNet-50 & 50M&  56.6 & 79.3 & 62.6 & 19.7 & 30.6 & 21.9\\
    WaterMask  & ResNet-101 & 67M & 59.0 & 80.6 & 67.2 & 38.7 & 54.9 & 43.2\\
    \hline
    \multicolumn{9}{c}{Transformer-Based Methods} \\
    \hline
    SAM+BBox & ViT-H  & 641M& 45.9 & 65.9 & 52.1 & 26.4 & 38.9 & 29.0 \\
    SAM+Mask  & ViT-H & 641M & 55.1 & 80.2 & 62.8 & 38.5 & 56.3 & 44.0 \\
    RSPrompter  & ViT-H & 632M & 58.2 & 79.9 & 65.9 & 40.2 & 55.3 & 44.8\\
    USIS-SAM  & ViT-H & 701M & \underline{59.7} & \underline{81.6} & \underline{67.7} & \underline{43.1} & \underline{59.0} &\underline{48.5} \\
    \textit{DiveSeg}(Ours) & ViT-L & 390M & \textbf{64.1} & \textbf{82.8} & \textbf{72.2} & \textbf{48.4} &\textbf{ 62.3} & \textbf{54.4} \\
    \hline  \hline
    \end{tabular}}
    \end{center}
    \vspace{-3mm}
    \caption{Quantitative comparisons with state-of-the-arts on the USIS10K dataset. The best results are highlighted in bold, while the second-best results are indicated with underlines.}
    \label{tab:USIS10K}
    \vspace{-3mm}
\end{table*}

\begin{figure*}[h]
    \vspace{2mm}
    \centering
    \includegraphics[width=450pt]{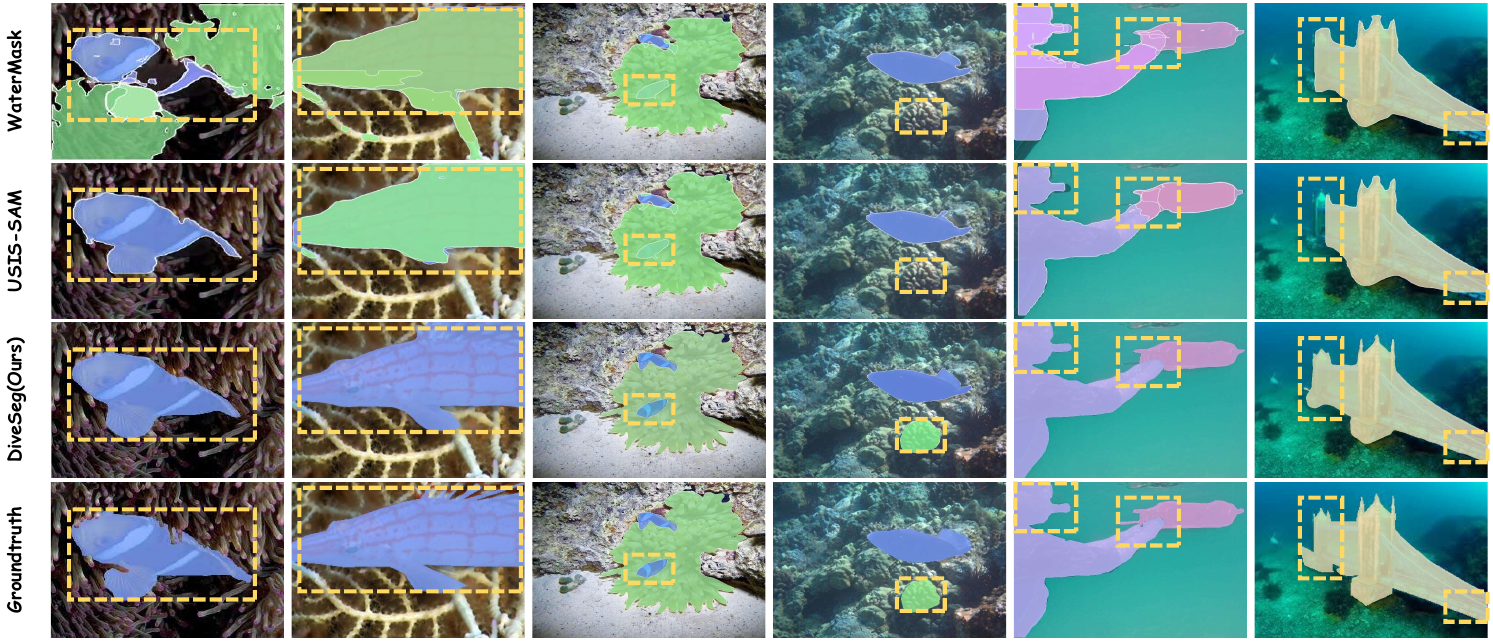} 
    \vspace{-2mm}
    \caption{Qualitative comparisons of \textit{DiveSeg} with SOTA UIS methods on the USIS10K and UIIS datasets.}
    \label{fig:Qualitative comparison}
    \vspace{-3mm}
\end{figure*}

\section{Experiments}

We conduct extensive experiments on UIIS \cite{lian2023watermask} and USIS10K \cite{lian2024diving} datasets to verify the effectiveness of the proposed \textit{DiveSeg}. Experimental results show that \textit{DiveSeg} achieves better performance than the existing SOTA methods. In addition, we also performed ablation experiments on the proposed \textit{AquaStyle Aligner} and \textit{ObjectPrior Prompter} to verify the effectiveness of individual modules.

\subsection{Experimental Setup}

\textbf{Datasets.}
The proposed method is evaluated on UIIS and USIS10K datasets. The UIIS dataset contains 3937 training images and 691 testing images, covering 7 instance categories, as follows: Fish, Reefs, Aquatic plants, Wrecks/Ruins, Human divers, Robots, and Sea-floor. The USIS10K dataset contains 10632 images, which are divided into training set, validation set and test set according to the ratio of 7:1.5:1.5. The dataset contains not only the seven instance categories mentioned above, but also class-agnostic labels to support class-agnostic instance detection tasks. 

\noindent \textbf{Evaluation Metrics.}
To ensure a fair and comprehensive comparison, we adopt the standard mask AP evaluation protocol commonly used in instance segmentation benchmarks. This includes metrics such as $mAP$, $AP_{50}$, and $AP_{75}$, which correspond to average precision computed at different IoU thresholds. These metrics collectively capture both coarse and fine-grained segmentation accuracy, allowing for a thorough assessment of the model's ability to detect and localize instance masks with varying levels of precision.



\noindent \textbf{Implementation Details.} 
We implemented the proposed approach using PyTorch and the Detectron2 framework. Training was conducted on an NVIDIA A100 GPU with a batch size of 8, using AdamW as the optimizer with a weight decay of 0.05. The initial learning rate was set to 1e-4, and the warm-up strategy was employed. The training lasted for 30,000 iterations, during which the learning rate was decayed to one-tenth of its initial value at the 23,000th and 27,000th iterations. The segmentation head adopted the Mask2Former~\cite{cheng2022masked} architecture. We applied classification loss and mask loss to constrain the predictions of Mask2Former, while binary cross-entropy (BCE) loss, IoU loss, and L1 loss were used to guide the learning of pseudo masks. 

\subsection{Comparison with State-of-the-arts}
\textbf{Quantitative Results.}
\textbf{(1) On the UIIS dataset.}
As shown in Table~\ref{tab:UIIS}, we compare our method against the SOTA algorithms, including Mask Scoring R-CNN~\cite{huang2019mask}, Cascade Mask R-CNN~\cite{cai2018cascade}, BMask R-CNN~\cite{cheng2020boundary}, Point Rend~\cite{li2019underwater}, $\mathrm{R}^3$-CNN~\cite{rossi2021recursively}, Mask Transfiner~\cite{ke2022mask}, Mask2Former, WaterMask and USIS-SAM~\cite{lian2024diving}. Our \textit{DiveSeg}, which employs ViT-L as the backbone, achieves $mAP$, $AP_{50}$, and $AP_{75}$ of 35.6, 52.0, and 38.5, respectively, representing improvements of 21.1\%, 15.6\%, and 19.2\% over USIS-SAM with a ViT-H backbone. Furthermore, \textit{DiveSeg} substantially outperforms WaterMask and other competitors across all metrics, demonstrating its superior capability in underwater instance segmentation.
\textbf{(2) On the USIS10K dataset.}
As summarized in Table 2, our approach demonstrates clear superiority over existing methods including S4Net~\cite{fan2019s4net}, RDPNet~\cite{wu2021regularized}, OQTR~\cite{pei2022transformer}, WaterMask, SAM+BBox~\cite{kirillov2023segment}, SAM+Mask~\cite{kirillov2023segment}, RSPrompter~\cite{chen2024rsprompter} and USIS-SAM. In the class-agnostic setting, it achieves $mAP$, $AP_{50}$, $AP_{75}$ of 64.1, 82.8, 72.2, outperforming USIS-SAM by 4.4, 1.2, and 4.5 percentage points, respectively. Under the multi-class setting, it attains $mAP$, $AP_{50}$, $AP_{75}$ of 48.4, 62.3, 54.4—substantially higher than USIS-SAM—while using only 55.6\% of its parameters. These results compellingly demonstrate that our model not only excels at class-agnostic instance segmentation but also delivers precise class-level and instance-level predictions in challenging underwater environments.

\noindent \textbf{Qualitative Results.}
As shown in Figure \ref{fig:Qualitative comparison}, we qualitatively compare our method with WaterMask and USIS-SAM on the UIIS and USIS10K test sets. Our approach more accurately delineates object boundaries for instance, successfully segmenting fish in shadows (Column 1) and underwater ruins (Column 6), and detecting coral missed by others (Column 4).
 In terms of classification accuracy, our model also outperforms the competitors. For example, in Column 2, other methods misclassify fish as coral due to complex environmental interference, whereas our model produces correct predictions.
Moreover, in cases of instance overlap such as fish and coral (Column 3) or divers with equipment (Column 5) our model maintains clear instance boundaries, while others often produce boundary confusion.

\begin{table}[!t]
    \small 
    \begin{center}
    \renewcommand{\arraystretch}{1.2}
    \setlength{\tabcolsep}{2mm}
    {\begin{tabular}{c|ccc}
    \hline\hline
    Methods &$mAP$ & $AP_{50}$ & $AP_{75}$ \\
    \hline
    Full Model & \textbf{35.6}& \textbf{52.0}& \textbf{38.5}\\
    w/o AquaStyle Aligner & 34.8 &50.6 &37.6 \\
    w/o ObjectPrior Prompter & 34.1  & 50.8 & 37.8 \\
    DINOv2+Mask2Former & 30.9  & 44.6 & 32.2 \\

    \hline\hline
    \end{tabular}}
    \end{center}
    \vspace{-3mm}
    \caption{Ablation study of individual components.}
    \label{tab:contribution}
    \vspace{-3mm}
\end{table}

\subsection{Ablation Study}

\noindent \textbf{Effectiveness of the Individual Modules.}
Table~\ref{tab:contribution} presents an ablation study evaluating the contributions of the \textit{AquaStyle Aligner} and the \textit{ObjectPrior Prompter}, based on a baseline that comprises a frozen DINOv2 backbone and a trainable Mask2Former decoder. Introducing the \textit{ AquaStyle Aligner} increases $mAP$ by 3.9 points, $AP_{50}$ by 6.0 points, and $AP_{75}$ by 5.4 points, while adding the \textit{ObjectPrior Prompter} yields improvements of 3.2, 6.2, and 5.6 points on these metrics, respectively. When both modules are combined, $mAP$ reaches 35.6, with corresponding further gains in $AP_{50}$ and $AP_{75}$. The full model achieves gains of 15.2\%, 16.6\%, and 19.6\% over the "DINOv2+Mask2Former" baseline, which demonstrates that the \textit{AquaStyle Aligner} and the \textit{ObjectPrior Prompter} play significant roles in adapting DINOv2 for underwater instance segmentation. 

\noindent \textbf{Ablation Study on the AquaStyle Aligner.}
Table~\ref{tab:ASA} reports an analysis of alternative adaptation strategies for DINOv2 to validate the efficacy of our \textit{AquaStyle Aligner}. "Frozen" denotes that all parameters of DINOv2 are kept fixed, and no additional learnable parameters are introduced in the encoder. First, we adopt the full fine-tuning strategy for DINOv2 (Row 2), which enables gradient updates for all parameters of the DINOv2 model that are kept frozen in the “Frozen” setting. However, this approach yields only marginal improvements over the frozen backbone, possibly due to catastrophic forgetting of the pre-trained knowledge caused by updating a large number of parameters. Moreover, inspired by recent studies demonstrating that low-rank adaptation (LoRA) and lightweight Adapters can effectively tailor pre-trained ViTs with minimal parameter overhead, we apply these techniques in Rows 3 and 4. Both approaches lead to substantial performance improvements, confirming the effectiveness of targeted parameter fine-tuning. The proposed \textit{AquaStyle Aligner} scheme (Row 5), which incorporates specialized underwater style extraction and effective style injection, surpasses these methods. This demonstrates that integrating underwater style features into DINOv2 provides more pronounced benefits for understanding and segmenting underwater scenes.

\begin{table}[!t]
    \small 
    \begin{center}
    \renewcommand{\arraystretch}{1.2}
    \setlength{\tabcolsep}{2.5mm}
    {\begin{tabular}{c|ccc}
    \hline\hline
    Methods & $mAP$ & $AP_{50}$ & $AP_{75}$ \\
    \hline
    Frozen & 30.9& 44.6 & 32.2\\
    Full Fine-tuning & 31.1 & 45.4 & 35.2 \\
    LoRA & 31.8 & 47.4& 34.6\\
    Adapter& 32.7 & 48.1 & 36.4\\
    AquaStyle Aligner &\textbf{34.1} &\textbf{50.8} &\textbf{37.8}  \\
    \hline\hline
    \end{tabular}}
    \end{center}
    \vspace{-3mm}
    \caption{Ablation study of \textit{AquaStyle Aligner}. }
    \vspace{-3mm}
    \label{tab:ASA}
\end{table}

\section{Conclusion}
This paper introduces \textit{DiveSeg}, a novel Underwater Instance Segmentation (UIS) framework built on the DINOv2 visual foundation model. It tackles two key challenges of UIS: underwater scene adaptation and underwater object adaptation. \textit{DiveSeg} has two core components: \textit{AquaStyle Aligner}, which embeds underwater color style features into DINOv2 fine-tuning via Fourier decomposition and cross-attention to boost domain adaptation; and \textit{ ObjectPrior Prompter}, which uses binary segmentation prompts to provide object-level priors, enabling instance segmentation through dual object- and instance-level reasoning.
Extensive experiments on UIIS and USIS10K datasets show \textit{DiveSeg} achieves state-of-the-art performance, significantly outperforming existing methods in both quantitative and qualitative results.

\section{Acknowledgments}
This work was supported in part by the National Natural Science Foundation of China under Grant 62471278, and in part by the Taishan Scholar Project of Shandong Province under Grant tsqn202306079.

\bigskip

\bibliography{aaai2026}

\end{document}